\pdfoutput=1

\documentclass[11pt]{article}

\usepackage{acl}

\usepackage{times}
\usepackage{latexsym}
\usepackage{graphicx}

\usepackage[T1]{fontenc}

\usepackage[utf8]{inputenc}

\usepackage{microtype}

\title{Exploring the Constructicon: \\
Linguistic Analysis of a Computational CxG}

\author{Jonathan Dunn \\
  Department of Linguistics and \\
  New Zealand Institute for Language, Brain and Behaviour \\
  University of Canterbury \\
  Christchurch, New Zealand \\
  \texttt{jonathan.dunn@canterbury.ac.nz} }

\begin{document}
\maketitle
\begin{abstract}
Recent work has formulated the task for computational construction grammar as producing a constructicon given a corpus of usage. Previous work has evaluated these unsupervised grammars using both internal metrics (for example, Minimum Description Length) and external metrics (for example, performance on a dialectology task). This paper instead takes a linguistic approach to evaluation, first learning a constructicon and then analyzing its contents from a linguistic perspective. This analysis shows that a learned constructicon can be divided into nine major types of constructions, of which \textit{Verbal} and \textit{Nominal} are the most common. The paper also shows that both the token and type frequency of constructions can be used to model variation across registers and dialects.
\end{abstract}

\section{Introduction}

Construction Grammar (CxG) is a usage-based approach to language which views grammatical structure as a set of form-meaning mappings called a \textit{constructicon} \cite{l08}. From this usage-based perspective, a \textit{construction} could belong in the grammar either (i) because it is sufficiently entrenched (i.e., frequent) that it is stored and processed as a unique item or (ii) because it is sufficiently irregular (i.e., idiomatic) that it requires a unique grammatical description \cite{g06a}. The advantage of CxG from this perspective is that it focuses on explaining the creativity, the flexibility, and the idiosyncrasy of actual language use in real-world settings \cite{Goldberg2019}.

Given this focus of CxG as a linguistic theory, the ideal computational implementation must be data-driven and unsupervised. For example, approaches which rely on manual annotations derived from individual introspection \cite{Steels2017} fail to capture the usage-based foundations of CxG, in addition to being unreproducible and difficult to scale. For this reason, most recent work on computational CxG has taken an unsupervised learning approach to forming constructicons \cite{d17, Dunn2022b}. Such an unsupervised approach has its own challenges, however, especially the challenge of evaluation. Grammars from other syntactic paradigms can be evaluated by annotating a gold-standard corpus and then measuring the ability of both supervised and unsupervised models to predict those same sets of annotations (cf., \citealt{zeman-etal-2017-conll, zeman-EtAl:2018:K18-2}). Given its usage-based foundations, this approach to evaluation is simply not feasible for computational CxG because the standard for what counts as a construction depends to some degree on the corpus or the community of speaker-hearers that is being observed.

For this reason, recent work on computational CxG has undertaken both internal and external evaluations for determining which one of a set of posited constructicons is better. An internal metric measures the fit between a grammar and a given corpus to determine which alternative constructicon offers a better description \cite{d18, Dunn2019}. This work has drawn on Minimum Description Length \cite{Goldsmith2001, Goldsmith2006} as an evaluation metric because it combines both descriptive adequacy (i.e., the fit between the grammar and the test set) and model complexity (i.e., the number and the type of constructions in the grammar).

An external metric evaluates and compares constructicons using their performance when applied to a specific prediction task. Recent work has focused on the use of computational CxG for modelling individual differences \cite{Dunn2021a}, register variation \cite{Dunn2021}, and population-based dialectal differences \cite{d18b, Dunn2019a, 10.3389/frai.2019.00015, dunn-wong-2022-stability}. Because CxG is a usage-based paradigm, the definition of a construction that is referenced above depends on both entrenchment and idiomaticity. Both of these are properties of a corpus of usage rather than properties of a language as a whole. In other words, it is only meaningful to describe \textit{entrenchment} relative to a particular individual, dialectal community, or context of production. These external tasks have therefore focused on the degree to which computational CxG can in fact account for differences in usage across these dimensions.

The contribution of this paper is to undertake a detailed qualitative and quantitative evaluation of a learned grammar. While it is not possible to start with gold-standard linguistic annotations of constructions, it is possible to apply a linguistic analysis to the output of an unsupervised, usage-based framework. We start by describing the model and the data which are used to learn the constructicon (Section 2) before presenting examples of types of constructions that it contains (Section 3). We then proceed to a quantitative analysis of the grammar (Section 4). Finally, we end with a discussion of the challenge of parsing a nested and hierarchical grammar which contains representations at different levels of abstraction (Section 5).

\section{Methods and Data}

Computational CxG is a theory in the form of a grammar induction algorithm that provides a reproducible constructicon given a corpus of exposure \cite{d17, Dunn2022b}. The theory is divided into three components, each of which models a particular aspect of the emergence of constructicons given exposure to a corpus of usage.

First, a psychologically-plausible measure of association, the $\Delta P$, is used to measure the entrenchment of potential constructions \cite{Ellis2007, DunnIJCL}. These potential constructions are sequences of lexical, syntactic, and semantic slot-constraints. The problem of \textit{category formation} is to define the inventory of fillers that are used for slot-constraints. In this implementation, lexical constraints are based on word-forms, without lemmatization. Syntactic constraints are formulated using the universal part-of-speech tagset \cite{pdm12} and implemented using the Ripple Down Rules algorithm \cite{nn}. Semantic constraints are based on distributional semantics, with k-means clustering used to discretize fastText embeddings \cite{Grave2018}. The semantic constraints in the examples in this paper are formulated using the index of the corresponding clusters, a simple notational convention.

Second, an association-based beam search is used to identify constructions of arbitrary length by finding the most entrenched representations in reference to a matrix of $\Delta P$ values \cite{Dunn2019}. The beam search parsing strategy allows the grammar to avoid relying on heuristic frames and templates for producing potential constructions.

Third, a measure of fit based on the Minimum Description Length paradigm is used to balance the increased storage of item-specific constructions against the increased computation of more generalized constructions \cite{d18}. The point is that any construction could become entrenched but more idiomatic constructions come at a higher cost.

The contribution of this paper is to evaluate this existing model of CxG \cite{Dunn2022b} rather than to alter its overall method of learning a constructicon. We therefore apply the model without further discussion of its implementation and focus instead on a linguistic analysis of the resulting constructicon. The data used to learn grammars is collected from three sets of corpora: social media (Twitter), non-fiction articles (Wikipedia), and web pages (from the Common Crawl) drawn from the \textit{Corpus of Global Language Use} \cite{Dunn2020}. This training corpus contains 2 million words per register for a total of 6 million words. 

From a usage-based perspective, exposure to language continues after the grammar has been acquired and such exposure might change the entrenchment of particular constructions. The model thus undertakes a second pruning stage which updates the constructicon given an additional 2 million words of exposure \cite{Dunn2022b}. The model observes sub-corpora from each of the three registers in increments of 100k words. Each construction in the grammar receives an activation weight with an initial value of 1. For each sub-corpus in which a construction is not observed, its weight decays by 0.25. For each sub-corpus in which a construction is observed, its weight is returned to 1. When a construction’s weight falls below 0, it is forgotten and removed from the grammar.

This is a simple model of the way in which continued exposure leads to the forgetting of previously entrenched constructions. While somewhat arbitrary, the decay rate (0.25) is chosen to ensure that a construction is not forgotten simply because it occurs primarily in a specific register: this decay rate means that a construction must be absent from four successive sub-corpora, thus ensuring that each of the three registers has been observed. Thus, this pruning method removes unproductive constructions given additional exposure while ensuring that all three registers remain represented. A package for reproducing this grammar induction algorithm is available\footnote{\href{https://github.com/jonathandunn/c2xg/releases/tag/v1.0}{https://www.github.com/jonathandunn/c2xg}} as well as the specific grammars used in this study.\footnote{\href{https://doi.org/10.18710/CES0L8}{https://doi.org/10.18710/CES0L8}}

This method produces a constructicon that contains 12,856 constructions. The analysis in this paper is based on using this constructicon to annotate samples of 1 million words from 12 independent corpora: Project Gutenberg \cite{Rae2019},
Wikipedia \cite{Ortman2018},  European Parliament proceedings \cite{Tiedemann2012a}, news article comments \cite{Kesarwani2018}, product reviews \cite{Zhang2015}, blogs \cite{Schler2006}, and tweets from six countries (with 1 million words representing each country; \citealt{Dunn2020}). This range of corpora allows us to consider both register (different contexts of production) and dialect (different populations using the same register) when measuring the frequency and the productivity of individual constructions in the grammar.

\section{Categorizing Constructions}

In this section we categorize the learned constructions to aid our quantitative analysis of the contents of the constructicon. We annotate a random sample of 20\% of the constructions using the categorization described below, thus allowing an estimate of the overall composition of the grammar. The primary categories are \textit{Verbal}, \textit{Nominal}, \textit{Adjectival}, \textit{Adpositional}, \textit{Transitional}, \textit{Clausal}, \textit{Adverbial}, \textit{Sentential}, and \textit{Fixed Idioms}. These categories are defined and exemplified in this section.

The first category consists of \textsc{verbal} constructions. As shown in (1), we notate the construction using its slot-constraints, with each slot separated by dashes. Lexical constraints are shown in italics; syntactic constraints are shown in small caps; and semantic constraints are shown using the index of their distributional cluster (e.g., <521>). Using this notation, the construction in (1) is a simple passive verb phrase in a continuous aspect, defined using primarily syntactic constraints.

\vspace*{0.2cm}\noindent (1) [ \textsc{aux} -- \textit{being} -- \textsc{verb} ] \\
\hspace*{0.4cm}(1a) were being proposed \\
\hspace*{0.4cm}(1b) was being spread \\
\hspace*{0.4cm}(1c) is being invaded \\
\hspace*{0.4cm}(1d) am being kept \\

The verbal construction in (2) now contains a semantic constraint (<521>). This domain contains lexical items like \textit{house} and \textit{carriage}, all locations that can be moved into or out of. The construction thus captures a meaning-based pattern of movement in relation to some area.

\vspace*{0.2cm}\noindent (2) [ \textsc{verb} -- \textsc{adp} -- \textsc{det} -- <521> ] \\
\hspace*{0.4cm}(2a) come to this house \\
\hspace*{0.4cm}(2b) leaped into a carriage \\
\hspace*{0.4cm}(2c) seated at that window \\
\hspace*{0.4cm}(2d) hurried across the room \\
\hspace*{0.4cm}(2e) lying on the floor \\

A lexical constraint for the main verb is shown in the construction in (3). This leads to an idiomatic usage of \textit{play}, a set of utterances whose behaviour differs from the basic transitive verb phrase. The construction in (4) shows the influence of a lexical constraint in a different position, here \textit{time} as a noun introducing the verb phrase. This again results in idiomatic utterances with behaviour more specific than a construction with only syntactic constraints. Finally, the lexical constraint in (5) defines a particle verb, again with idiomatic semantics resulting for the utterances in (5a) through (5e). This series of examples shows how a lexical constraint in different locations within a verb phrase leads to different types of idiomatic verbal constructions.

\vspace*{0.2cm}\noindent (3) [ \textit{play} -- \textsc{det} -- \textsc{noun} ] \\
\hspace*{0.4cm}(3a) play the game \\
\hspace*{0.4cm}(3b) play the part \\
\hspace*{0.4cm}(3c) play the coquette \\
\hspace*{0.4cm}(3d) play the king \\
\\
\noindent (4) [ \textit{time} -- \textit{to} -- \textsc{verb} ] \\
\hspace*{0.4cm}(4a) time to plead \\
\hspace*{0.4cm}(4b) time to write \\
\hspace*{0.4cm}(4c) time to tell \\
\hspace*{0.4cm}(4d) time to consider \\
\hspace*{0.4cm}(4e) time to worry \\
\\
\noindent (5) [ \textit{to} -- \textsc{verb} -- \textit{down} ] \\
\hspace*{0.4cm}(5a) to sit down \\
\hspace*{0.4cm}(5b) to put down \\
\hspace*{0.4cm}(5c) to settle down \\
\hspace*{0.4cm}(5d) to bring down \\
\hspace*{0.4cm}(5e) to strike down \\

While these examples are relatively simple verbal constructions, a more complex example is shown in (6). This construction contains a main verb with an infinitive complement followed by an argument that takes the form of a noun phrase. The entrenchment of these more complex constructions shows the flexibility of computational CxG as well as the infeasibility of relying on the introspection of individual linguists.

\vspace*{0.2cm}\noindent (6) [\textsc{verb} -- \textit{to} -- \textit{be} -- <830> -- \textsc{adp} -- \textsc{det} -- \textsc{noun}] \\
\hspace*{0.4cm}(6a) seem to be unaware of the fact \\
\hspace*{0.4cm}(6b) came to be known as the \textit{Newcastle} \\
\hspace*{0.4cm}(6c) have to be supplied from that source \\
\hspace*{0.4cm}(6d) is to be found in the world \\
\hspace*{0.4cm}(6e) expect to be ushered into the temple \\

Moving to \textsc{nominal} constructions, the first examples show the influence that a semantic constraint in one slot exerts across the entire construction. We focus here on complex nominal constructions, with both of these first examples containing a subordinate adpositional phrase within the noun phrase. In each case, the noun in the adpositional phrase is constrained to a specific semantic domain. In (7), this leads to lexical items like \textit{empire} and \textit{palace} and, in (8), like \textit{ground} and \textit{road}. Not all examples of a construction are perfect matches; an example of this is shown in (8e), marked with an asterisk, in which the first word is actually a mistagged verb rather than a noun.

\vspace*{0.2cm}\noindent (7) [ \textsc{noun} -- \textit{of} -- \textsc{det} -- <587> ] \\
\hspace*{0.4cm}(7a) part of the empire \\
\hspace*{0.4cm}(7b) inmates of the palace \\
\hspace*{0.4cm}(7c) guardianship of the wanderer \\
\hspace*{0.4cm}(7d) pursuit of a chimera \\
\hspace*{0.4cm}(7e) circuit of the citadel \\
\\
\noindent (8) [ \textsc{noun} -- \textsc{adp} -- \textit{the} -- <484> ] \\
\hspace*{0.4cm}(8a) feet on the ground \\
\hspace*{0.4cm}(8b) side of the road \\
\hspace*{0.4cm}(8c) law of the land \\
\hspace*{0.4cm}(8d) entrance of the path \\
\hspace*{0.4cm}(8e) journey through the forest \\
\hspace*{0.4cm}(8e) *wanders around the forest \\
\\
\vspace*{0.2cm}\noindent (9) [ \textit{one} -- \textsc{adp} -- \textit{the} -- \textit{best} -- \textsc{noun} ] \\
\hspace*{0.4cm}(9a) one of the best paintings \\
\hspace*{0.4cm}(9b) one of the best apologies \\
\hspace*{0.4cm}(9c) one of the best examples \\
\hspace*{0.4cm}(9d) one of the best books \\
~

More idiomatic noun phrases, with lexical constraints, are shown in (9) and (10). In the first, an adpositional phrase \textit{one of the best} functions as a single adjective. In the second, a superlative adjective frames the core noun phrase. In both cases, these constructions provide additional flexibility to describe unique nominal phrases, made into constructions by their entrenchment and their idiosyncrasy in this set of usage.

~

\noindent (10) [ \textit{the} -- \textit{most} -- \textsc{adj} -- \textsc{noun} ] \\
\hspace*{0.4cm}(10a) the most amusing instance \\
\hspace*{0.4cm}(10b) the most violent writhings \\
\hspace*{0.4cm}(10c) the most astounding instances \\
\hspace*{0.4cm}(10d) the most important generalizations \\
\hspace*{0.4cm}(10e) the most unfavourable circumstances \\

A single example of an \textsc{adjectival} construction is shown in (11). While the previous nominal constructions included adjectival material within them, this construction as a whole provides a modifier for a noun phrase. For example, (11e) as an abstract adjective could be combined with a variety of nouns like \textit{immigrants}, \textit{the elderly}, or \textit{house sparrows} to form a larger nominal construction.

\vspace*{0.2cm}\noindent (11) [ \textit{huge} -- \textsc{noun} -- \textit{of} ] \\
\hspace*{0.4cm}(11a) huge pair of \\
\hspace*{0.4cm}(11b) huge influx of \\
\hspace*{0.4cm}(11c) huge clumps of \\
\hspace*{0.4cm}(11d) huge piece of \\
\hspace*{0.4cm}(11e) huge population of \\

The next category is \textsc{adpositional} constructions, as shown in (12) through (14). As before, a semantic constraint leads to a meaning-based group of utterances, as with the terms specific to legal language in (12). In other words, this adpositional construction is specific to the category of nouns contained within it. A potentially problematic case is shown in (12e), here with what is likely a fixed idiom, where \textit{case} is not used in the legal sense. A lexical constraint for the head noun in (13) leads to idiosyncratic adpositional phrases with \textit{beginning}. Other adpositional constructions are more syntactically complex. For example, the phrase in (14) transitions from a noun into a relative clause which describes that noun.

\vspace*{0.2cm}\noindent (12) [ \textsc{adp} -- \textsc{det} -- <959> ] \\
\hspace*{0.4cm}(12a) in the case \\
\hspace*{0.4cm}(12b) of the provisions \\
\hspace*{0.4cm}(12c) as a rule \\
\hspace*{0.4cm}(12d) from the petitioners \\
\hspace*{0.4cm}(12e) ?in which case\\
\noindent (13) [ \textsc{adp} -- \textit{the} -- \textit{beginning} ] \\
\hspace*{0.4cm}(13a) towards the beginning \\
\hspace*{0.4cm}(13b) at the beginning \\
\hspace*{0.4cm}(13c) from the beginning \\
\hspace*{0.4cm}(13d) in the beginning \\
\hspace*{0.4cm}(13e) for the beginning \\
\\
\noindent (14) [ \textsc{adp} -- \textit{the} -- \textsc{noun} -- \textit{where} ] \\
\hspace*{0.4cm}(14a) in the world where \\
\hspace*{0.4cm}(14b) at the spot where \\
\hspace*{0.4cm}(14c) from the point where \\
\hspace*{0.4cm}(14d) near the ceiling where \\
~

The example of an adpositional phrase that transitions into a relative clause in (14) introduces another category of constructions, those which capture \textsc{transitional} material connecting other types of constructions. In particular, the constructions in this category capture different types of transitions without containing the substance of the involved structures themselves. For example, in (15) there is the introduction of a new main clause with a first-person verb phrase. In (16) there is the introduction of a subordinate clause. In (17) there is a comparison between two nominal constructions. The final example in (17e) represents a problematic parse: the phrase is likely \textit{at least} rather than \textit{least} alone. These examples show how this category serves to link other constructions together.

\vspace*{0.2cm}\noindent (15) [ \textit{but} -- \textit{i} -- \textsc{verb} ] \\
\hspace*{0.4cm}(15a) but i think \\
\hspace*{0.4cm}(15b) but i knew \\
\hspace*{0.4cm}(15c) but i regret \\
\hspace*{0.4cm}(15d) but i noticed \\
\\
\noindent (16) [ \textsc{sconj} -- \textsc{verb} -- \textit{to} ] \\
\hspace*{0.4cm}(16a) without seeming to \\
\hspace*{0.4cm}(16b) because according to \\
\hspace*{0.4cm}(16c) as opposed to \\
\hspace*{0.4cm}(16d) while listening to \\
\hspace*{0.4cm}(16e) in resorting to \\
\\
\noindent (17) [ \textsc{adv} -- <917> -- \textit{than} ] \\
\hspace*{0.4cm}(17a) far deeper than \\
\hspace*{0.4cm}(17b) considerably better than \\
\hspace*{0.4cm}(17c) now more than \\
\hspace*{0.4cm}(17d) much smaller than \\
\hspace*{0.4cm}(17e) *least better than \\

While transitional constructions focus mainly on the connecting element, \textsc{clausal} constructions are those which contain a significant portion of a subordinate clause. For instance, (18) is an example of a relative clause embedded within a larger noun phrase and (19) of a relative clause in which the subject is defined by the proceeding element. A problematic example is shown in (19e), where the phrase \textit{a lot} is treated as two separate slots. The complex subordinate clause in (20) consists of a gerund within an adpositional phrase, where the verb is further defined by a semantic constraint. Finally, a reduced relative clause is captured by (21), again with a semantic constraint on the verb. This series of examples shows the way in which subordinate clauses are captured in the grammar.

\begin{figure*}[t]
\centering
\includegraphics[width = 450pt]{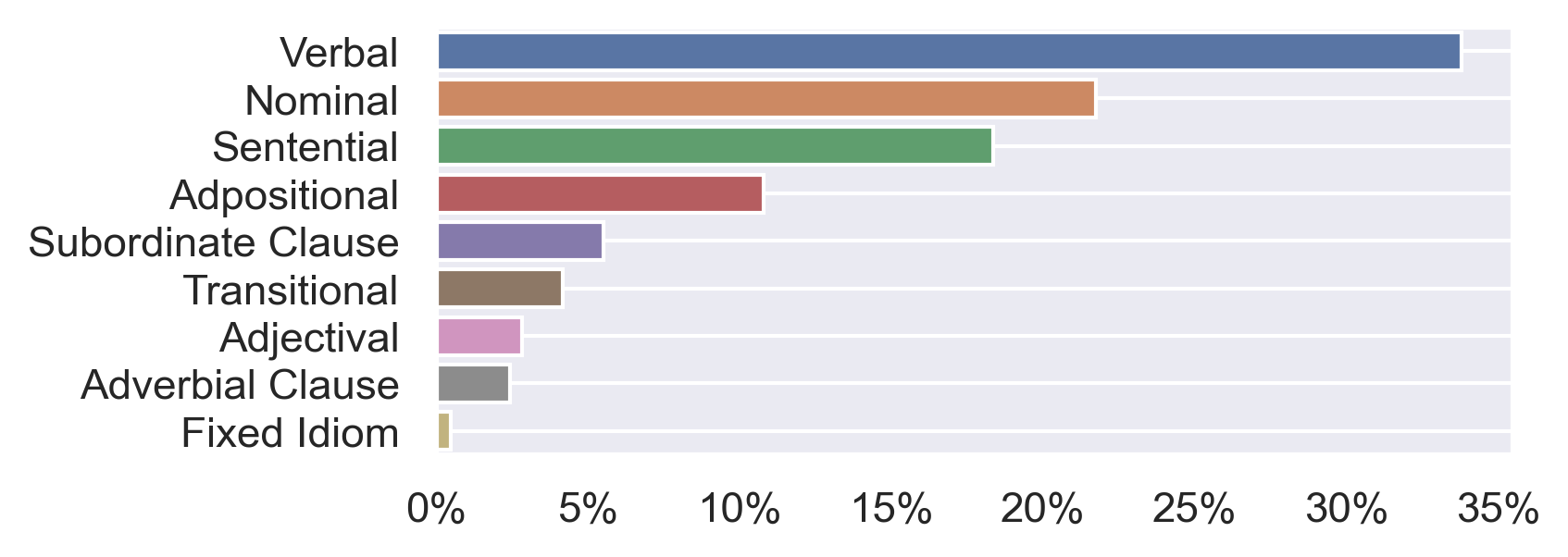}
\caption{Distribution of Construction Types in the Grammar}
\label{fig1}
\end{figure*}

~

\noindent (18) [ \textsc{noun} -- \textsc{adp} -- \textit{those} -- \textit{who} ] \\
\hspace*{0.4cm}(18a) hearts of those who \\
\hspace*{0.4cm}(18b) arguments of those who \\
\hspace*{0.4cm}(18c) side of those who \\
\hspace*{0.4cm}(18d) minds of those who \\
\hspace*{0.4cm}(18e) tactics of those who \\
\\
\noindent (19) [ \textit{which} -- \textsc{verb} -- \textit{a} -- \textsc{noun} ] \\
\hspace*{0.4cm}(19a) which formed a snare \\
\hspace*{0.4cm}(19b) which occasioned a detour \\
\hspace*{0.4cm}(19c) which presented a problem \\
\hspace*{0.4cm}(19d) which contained a letter \\
\hspace*{0.4cm}(19e) ? which looked a lot \\
\\
\noindent (20) [\textsc{sconj} -- <113> -- \textsc{det} -- \textsc{noun} -- \textit{of}] \\
\hspace*{0.4cm}(20a) by taking the life of \\
\hspace*{0.4cm}(20b) in sacrificing the rights of \\
\hspace*{0.4cm}(20c) after collecting the remains of \\
\hspace*{0.4cm}(20d) by applying a drop of \\
\hspace*{0.4cm}(20e) in neglecting the cultivation of \\
\\
\noindent (21) [ \textsc{det} -- \textsc{noun} -- \textit{he} -- <830> ] \\
\hspace*{0.4cm}(21a) the loan he solicited \\
\hspace*{0.4cm}(21b) the temple he discovered \\
\hspace*{0.4cm}(21c) the words he used \\
\hspace*{0.4cm}(21d) the life he led \\
\hspace*{0.4cm}(21e) the flask he carried \\
		
While these clausal constructions are connected into the main clause itself, the category of \textsc{adverbial} constructions contain clauses which are more independent of the structure of the main clause. For example, in (22) there is a gerund clause within an adpositional phrase, now with a semantic constraint. In (23) there is an adposition introducing a finite verb. And in (24), with a lexical constraint, there is a similar construction again with a finite verb. While similar to the clausal category, this class of constructions is less integrated with the main clause structure.

\vspace*{0.2cm}\noindent (22) [ \textsc{sconj} -- \textsc{verb} -- \textsc{adp} -- \textsc{det} -- <512> ] \\
\hspace*{0.4cm}(22a) in dealing with that section \\
\hspace*{0.4cm}(22b) after referring to the matter \\
\hspace*{0.4cm}(22c) as bearing on the question \\
\hspace*{0.4cm}(22d) without glancing within the volume \\
\hspace*{0.4cm}(22e) by bringing up the subject \\
		\\
\noindent (23) [\textsc{sconj} -- \textsc{pron} -- \textsc{aux} -- \textsc{verb} -- \textit{to}] \\ 
\hspace*{0.4cm}(23a) that it would come to \\
\hspace*{0.4cm}(23b) if he had lived to \\
\hspace*{0.4cm}(23c) as they were trying to \\
\\
\noindent (24) [ \textit{when} -- \textsc{det} -- \textsc{noun} -- \textit{is} ] \\
\hspace*{0.4cm}(24a) when the end is \\
\hspace*{0.4cm}(24b) when a man is \\
\hspace*{0.4cm}(24c) when the heart is \\
\hspace*{0.4cm}(24d) when the patient is \\
\hspace*{0.4cm}(24e) when the temperature is \\
\\
\vspace*{0.2cm}\noindent (25) [ \textsc{pron} -- \textit{were} -- \textsc{verb} -- \textsc{adp} ] \\
\hspace*{0.4cm}(25a) we were accosted by \\
\hspace*{0.4cm}(25b) they were employed by \\
\hspace*{0.4cm}(25c) these were succeeded by \\
\hspace*{0.4cm}(25d) they were drilled by \\
\hspace*{0.4cm}(25e) ? who were barred from \\

~
\textsc{sentential} constructions contain the structure of the main clause. This category overlaps to some degree with verbal constructions; the key difference is that the sentential constructions contain the subject while verbal constructions do not. A simple passive clause is shown in (25), together with an adpositional argument. In many examples, this adpositional argument specifies the agent, but the example in (25e) differs in specifying a location. An active clause introducing an indirect speech clause is shown in (26), constrained to the subject \textit{he}. Finally, a sequence of main verb and infinitive is shown in (27), with the final verb defined using a semantic constraint.

~

\noindent(26) [ \textit{he} -- \textsc{verb} -- \textit{that} ] \\
\hspace*{0.4cm}(26a) he remembered that \\
\hspace*{0.4cm}(26b) he said that \\
\hspace*{0.4cm}(26c) he realised that \\
\hspace*{0.4cm}(26d) he discovered that \\
\hspace*{0.4cm}(26e) he promised that \\
\\
(27) [ \textit{they} -- \textsc{verb} -- \textsc{part} -- <583> ] \\
\hspace*{0.4cm}(27a) they began to draw \\
\hspace*{0.4cm}(27b) they threatened to destroy \\
\hspace*{0.4cm}(27c) they chose to assert \\
\hspace*{0.4cm}(27d) they wanted to persuade \\
\hspace*{0.4cm}(27e) they began to look \\

\begin{table*}[t]
\centering
\begin{tabular}{|r|cc|cc|cc|cc|cc|cc|}
\hline
~ & \multicolumn{2}{|c|}{\textbf{Blogs}} & \multicolumn{2}{|c|}{\textbf{Comments}} & \multicolumn{2}{|c|}{\textbf{Parliament}} & \multicolumn{2}{|c|}{\textbf{Gutenberg}} & \multicolumn{2}{|c|}{\textbf{Reviews}} & \multicolumn{2}{|c|}{\textbf{Wikipedia}} \\
~ & \textit{Freq} & \textit{Type} & \textit{Freq} & \textit{Type} & \textit{Freq} & \textit{Type} & \textit{Freq} & \textit{Type} & \textit{Freq} & \textit{Type} & \textit{Freq} & \textit{Type} \\
\hline
\textit{Adjectival} & 57 & 36 & 69 & 43 & 66 & 40 & 79 & 59 & 80 & 45 & 73 & 43 \\
\textit{Adpositional} & 207 & 141 & 222 & 150 & 433 & 215 & 401 & 272 & 221 & 145 & 327 & 181 \\
\textit{Adverbial} & 118 & 87 & 107 & 80 & 117 & 79 & 95 & 80 & 127 & 88 & 56 & 45 \\
\textit{Idiom} & 32 & 3 & 33 & 2 & 54 & 13 & 12 & 4 & 27 & 3 & 13 & 2 \\
\textit{Nominal} & 95 & 82 & 128 & 109 & 261 & 184 & 189 & 163 & 123 & 101 & 179 & 138 \\
\textit{Sentential} & 199 & 115 & 144 & 103 & 176 & 107 & 144 & 110 & 195 & 111 & 109 & 77 \\
\textit{Clausal} & 156 & 99 & 157 & 112 & 182 & 117 & 154 & 112 & 152 & 97 & 70 & 58 \\
\textit{Transitional} & 102 & 75 & 96 & 77 & 103 & 72 & 107 & 89 & 108 & 82 & 49 & 43 \\
\textit{Verbal} & 137 & 104 & 143 & 116 & 188 & 142 & 139 & 122 & 144 & 108 & 116 & 86 \\
\hline
  \end{tabular}
  \caption{Mean Frequency and Productivity of Constructions by Category and Register}
  \label{tab:1}
\end{table*}

A more complex passive construction is shown in (28), containing both a semantic constraint on the main verb as well as an adpositional argument. Finally, a main clause with an existential \textit{there} as subject is shown in (29). As with the clausal constructions, these sentential constructions overlap with verbal constructions, thus illustrating the problem of parsing as clipping (c.f., Section 5).

\vspace*{0.2cm}\noindent (28) [ \textsc{noun} -- \textit{are} -- \textsc{adv} -- <830> -- \textsc{adp} ] \\
\hspace*{0.4cm}(28a) villages are thickly scattered about \\
\hspace*{0.4cm}(28b) recruits are never measured for \\
\hspace*{0.4cm}(28c) substances are universally regarded as \\
\hspace*{0.4cm}(28d) lines are then drawn from \\
\noindent (29) [ \textit{there} -- \textsc{verb} -- \textit{a} -- \textsc{noun} -- \textsc{adp} ] \\
\hspace*{0.4cm}(29a) there was a kind of \\
\hspace*{0.4cm}(29b) there is a habit of \\
\hspace*{0.4cm}(29c) there were a number of \\
\hspace*{0.4cm}(29d) there were a couple of \\
\hspace*{0.4cm}(29e) there came a sort of \\

The final category of constructions are \textsc{fixed idioms}, which here are mainly lexical constructions. These have a very limited number of types for each construction because the constraints are lexical: \textit{in favor of}, \textit{seems to be}, \textit{all the best}, or \textit{no matter \textsc{adv}}. Taken together, the categories illustrated in this section describe the contents of the learned constructicon. A quantitative analysis of the distribution of construction types and their properties follows in the next section.

\subsection{Marginal Examples of Categories}

Not all constructions that are classified as belonging to a given category are equally good examples of that category. This section provides a few examples of such marginal tokens in order to provide a more transparent picture of the grammar as a whole. Starting with a construction categorized as adjectival in (30), we could also see this being categorized as a nominal construction. The reason behind this annotation decision is that the overall unit is used to describe a part of some piece of writing.

~

\noindent (30) [ \textit{beginning} -- \textsc{adp} -- \textsc{det} -- \textsc{noun} ] \\
\hspace*{0.4cm}(30a) beginning of this note \\
\hspace*{0.4cm}(30b) beginning of the article \\

A marginal example of a nominal construction is shown in (31). Here, this sequence of noun and adpositional phrase, when taken in context, is quite likely to be two separate arguments of a double object verb phrase: for example, "They [ran [this country] [with the help...]]. However, the construction itself only includes the two arguments on their own. At the same time, (31) would clip together nicely with a verbal construction (c.f., Section 5).

~

\noindent(31) [\textit{this} -- \textsc{noun} -- \textsc{adp} -- \textit{the} -- \textsc{noun}] \\
\hspace*{0.4cm}(31a) this country with the help \\
\hspace*{0.4cm}(31b) this morning to the surprise \\
~\\
\noindent (32) [ \textsc{verb} -- \textit{by} -- \textsc{det} -- <88> ] \\
\hspace*{0.4cm}(32a) occcupied by a foreign \\
\hspace*{0.4cm}(32b) used by the american \\
~

A final marginal example is shown in (32), here within the verbal category. This example is a passive verb together with a prepositional phrase that expresses the agent. The issue here is that only part of the noun phrase specifying the agent is explicitly defined, and the slot constraint is semantic. From the perspective of clipping constructions, many noun phrases could be merged here but would not experience the same emergent relationships between slot-constraints. In other words, the impact of the semantic constraint would not transcend the construction boundary. These examples are meant to show some weaknesses of both the categorization scheme and the constructions themselves.
		
\section{Distribution of Construction Types}

\begin{figure*}[t]
\centering
\includegraphics[width = 450pt]{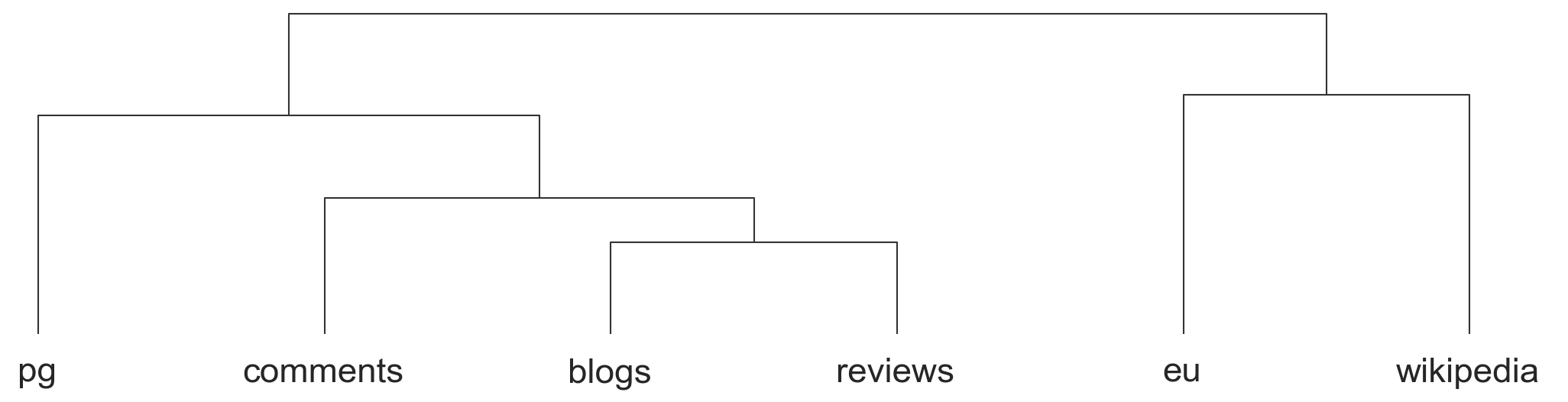}
\includegraphics[width = 450pt]{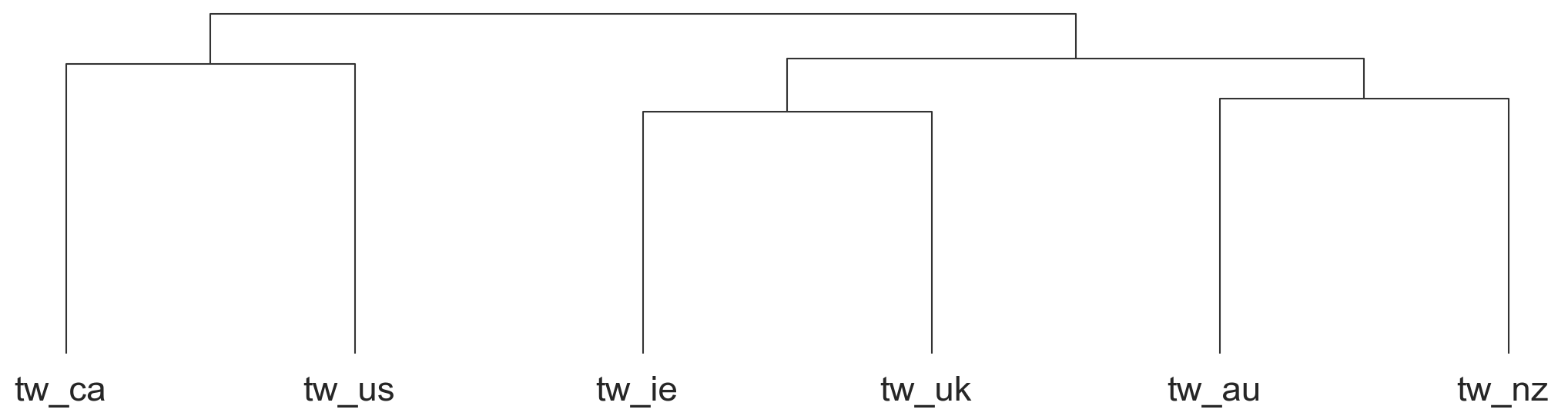}
\caption{Clustering of Corpora Using Burrow's Delta, Register (Above) and Dialect (Below)}
\label{fig2}
\end{figure*}

The first step in quantifying the contents of the constructicon is to estimate the relative distribution across these nine categories. This is shown in Figure \ref{fig1} using annotations of 20\% of the grammar to estimate the overall distribution. The y-axis contains a bar chart for each category of construction and the x-axis shows the percent of the constructicon which falls into that category.

Thus, for example, the most frequent type of construction is \textit{verbal} at 33.7\% of the grammar, followed by \textit{nominal} at 21.7\% and \textit{sentential} at 18.3\%. This distribution is not surprising given that verbs and nouns are the most common open-class lexical items and that sentential clauses form the basic structure of sentences.

The next step is to measure the frequency of each construction and the number of its unique types, thus capturing its productivity. These measures of frequency and productivity are corpus-specific in the sense that different constructions are more likely to be used in specific contexts or by specific populations. We thus consider 12 distinct corpora of 1 million words each, six representing distinct registers and six representing distinct populations within the same register.

Starting with a comparison across registers, Table \ref{tab:1} shows the mean frequency of tokens and the mean number of types for each class of constructions in each register-specific corpus. For example, the Project Gutenberg corpus has significantly more types per adpositional construction than the corpus of blogs. While some categories of construction are more common in the grammar, the measures in Table \ref{tab:1} take the average for each category. While there are more verbal constructions in the grammar, for example, adpositional and sentential constructions have more tokens per construction.

The frequency of each category of construction (i.e., the mean number of tokens) also provides a view of the grammatical differences between these six registers. For instance, blogs contain fewer adpositional constructions than other registers while published books and speeches in parliament contain approximately twice as many overall. Wikipedia articles contain many fewer cases of clasual and transitional constructions, indicating a register with fewer embedded clauses. Further, blogs have nearly twice as many sentential constructions (i.e., base main clauses) as Wikipedia, but many fewer adpositional phrases. This would indicate that information can be packaged in short sentences or in additional adpositional constructions, depending on the register. Note that another set of Wikipedia corpora was available during the grammar learning process, so that the reduced frequencies of these types are not simply a matter of under-fitting the register.

The next question is whether the differences in frequency of individual constructions across corpora are random or whether they reveal underlying relationships between the corpora themselves. In other words, given the frequencies of each construction in the grammar, we would expect a meaningful grammar to create meaningful relationships between conditions. A \textit{condition} in this case refers to the register or the population represented by the corpus. This is shown in Figure \ref{fig2} using Burrow's Delta to calculate the distances between corpora and then hierarchical clustering to visualize relationships based on these distances.

The figure shows relationships between registers on the top. The two core clusters are with modern formal documents (\textsc{eu} and \textsc{wikipedia}) and digital crowd-sourced documents (\textsc{comments} and \textsc{blogs} and \textsc{reviews}). The books from Project Gutenberg, from a different historical period, are an outlier. On the bottom the figure shows relationships between different dialects within the same register (tweets). The core pairs are the countries which are closest in geographic terms: Ireland and the UK together with Australia and New Zealand, with Canada and the US as a distant pair. In both cases, we see that the frequencies of constructions in the grammar provide meaningful relationships between both registers and dialects. This is important because it shows that the differing frequencies of constructions are not simply arbitrary patterns from this particular model but also reproduce two sets of real-world relationships.

\section{Clipping: The Problem of Parsing}

The analysis in this paper has categorized and described the kinds of constructions that are contained in a learned constructicon, has quantified the frequency and productivity of each kind, and has shown that the usage of these constructions can reconstruct meaningful relationships between corpora. The analysis of construction types in Section 3, however, reveals a major challenge in this approach to computational CxG: the unification or \textit{clipping together} of these constructions into complete utterances during parsing \cite{10.1093/oxfordhb/9780195396683.013.0005}.

The idea in CxG is that word-forms are not the basic building blocks of grammar. Rather, the types of constructions analyzed in this paper form the basic units, themselves built out of slot-constraints that depend on basic category formation processes. With the exception of short utterances, however, no single construction provides a complete description of a linguistic form. These constructions must be clipped together: a sentential construction, for example, joined with a verbal construction and then a nominal construction. CxG posits a continuum between the lexicon and the grammar, so that the constructicon contains basic units at different levels of abstraction. We must distinguish, however, between \textbf{first-order constructions} of the type discussed in this paper and \textbf{second-order constructions} which are formed by clipping together these lower constructions. A complete constructicon would thus also contain emergent structures formed from multiple first-order constructions. 

As a desideratum for future developments, we can conceptualize two types of second-order constructions: First, \textsc{slot-recursion} would allow a higher-order construction to contain first-order constructions as slot-fillers. For example, the set of sentential constructions could be expanded by allowing verbal constructions to fill verbal slots. Second, \textsc{slot-clipping} would allow two overlapping constructions to be merged, for instance connecting a transitional construction with a verbal construction. An overlapping shared slot-constraint would license such slot-clipping unifications.

\section{Conclusions}

The main contribution of this paper has been to provide a qualitative linguistic analysis of a learned construction grammar, providing a new perspective on grammars which have previously been evaluated from a quantitative perspective. We presented a division of construction types into nine categories such as \textit{Verbal} and \textit{Nominal}, with those two open-class categories the most common. The discussion of examples shows both the range and the robustness of computational construction grammar.

This linguistic analysis does point to two current weaknesses: First, not all constructions fit nicely into the categories used for annotation (c.f., Section 3.1). A truly usage-based grammar does not necessarily align with introspection-based analysis, especially in regards to boundaries between constructions. Introspection often focuses on constructions which are complete or self-contained units, while the computational constructions place common pivot points at boundaries. Second, these constructions do not generally describe entire utterances, so that we must consider a form of clipping to provide complete parses (c.f., Section 5).

From a quantitative perspective, the analysis of register and dialectal differences shows that the productivity of these constructions also reproduces expected relationships between corpora. This is important for providing an external evaluation of the grammar: the differences between registers, for example, show how functions which are salient in a given communicative situation ultimately drive constructional frequencies. In other words, the frequencies of different types of constructions reflect meaningful patterns in real-world usage.

\bibliography{References}

\begin{thebibliography}{31}
\expandafter\ifx\csname natexlab\endcsname\relax\def\natexlab#1{#1}\fi

\bibitem[{Dunn(2017)}]{d17}
J~Dunn. 2017.
\newblock \href {https://doi.org/https://doi.org/10.1017/langcog.2016.7}
  {{Computational Learning of Construction Grammars}}.
\newblock \emph{Language \& Cognition}, 9(2):254--292.

\bibitem[{Dunn(2018{\natexlab{a}})}]{d18b}
J.~Dunn. 2018{\natexlab{a}}.
\newblock \href {https://doi.org/https://doi.org/10.1515/cog-2017-0029}
  {{Finding Variants for Construction-Based Dialectometry: A Corpus-Based
  Approach to Regional {CxGs}}}.
\newblock \emph{Cognitive Linguistics}, 29(2):275--311.

\bibitem[{Dunn(2018{\natexlab{b}})}]{d18}
J~Dunn. 2018{\natexlab{b}}.
\newblock \href {https://doi.org/http://dx.doi.org/10.7275/R59P2ZTB} {{Modeling
  the Complexity and Descriptive Adequacy of Construction Grammars}}.
\newblock \emph{In Proceedings of the Society for Computation in Linguistics},
  pages 81--90.

\bibitem[{Dunn(2018{\natexlab{c}})}]{DunnIJCL}
J~Dunn. 2018{\natexlab{c}}.
\newblock \href {https://doi.org/10.1075/ijcl.16098.dun} {{Multi-Unit
  Association Measures: Moving beyond pairs of words}}.
\newblock \emph{International Journal of Corpus Linguistics}, 23(2):183--215.

\bibitem[{Dunn(2019{\natexlab{a}})}]{Dunn2019}
J.~Dunn. 2019{\natexlab{a}}.
\newblock \href {https://doi.org/http://dx.doi.org/10.18653/v1/W19-2913}
  {{Frequency vs. Association for Constraint Selection in Usage-Based
  Construction Grammar}}.
\newblock In \emph{Proceedings of the Workshop on Cognitive Modeling and
  Computational Linguistics}, page 117–128.

\bibitem[{Dunn(2019{\natexlab{b}})}]{10.3389/frai.2019.00015}
J.~Dunn. 2019{\natexlab{b}}.
\newblock \href {https://doi.org/10.3389/frai.2019.00015} {{Global Syntactic
  Variation in Seven Languages: Toward a Computational Dialectology}}.
\newblock \emph{Frontiers in Artificial Intelligence}, 2:15.

\bibitem[{Dunn(2019{\natexlab{c}})}]{Dunn2019a}
J.~Dunn. 2019{\natexlab{c}}.
\newblock \href {https://doi.org/http://dx.doi.org/10.18653/v1/W19-1405}
  {{Modeling Global Syntactic Variation in English Using Dialect
  Classification}}.
\newblock In \emph{Proceedings of the Sixth Workshop on NLP for Similar
  Languages, Varieties and Dialects}, pages 42--53.

\bibitem[{Dunn(2020)}]{Dunn2020}
J.~Dunn. 2020.
\newblock \href {https://doi.org/10.1007/s10579-020-09489-2} {{Mapping
  Languages: the Corpus of Global Language Use}}.
\newblock \emph{Language Resources and Evaluation}, 54:999--1018.

\bibitem[{Dunn(2022)}]{Dunn2022b}
J.~Dunn. 2022.
\newblock \href {https://doi.org/10.1515/cog-2021-0106} {{Exposure and
  Emergence in Usage-Based Grammar: Computational Experiments in 35
  Languages}}.
\newblock \emph{Cognitive Linguistics}, 33:659--699.

\bibitem[{Dunn and Nini(2021)}]{Dunn2021a}
J~Dunn and A~Nini. 2021.
\newblock \href {https://www.aclweb.org/anthology/2021.cmcl-1.19.pdf}
  {{Production vs Perception: The Role of Individuality in Usage-Based Grammar
  Induction}}.
\newblock In \emph{Proceedings of the Workshop on Cognitive Modeling and
  Computational Linguistics}, pages 149--159.

\bibitem[{Dunn and {Tayyar Madabushi}(2021)}]{Dunn2021}
J.~Dunn and H~{Tayyar Madabushi}. 2021.
\newblock \href {https://doi.org/http://dx.doi.org/10.18653/v1/2021.conll-1.21}
  {{Learned Construction Grammars Converge Across Registers Given Increased
  Exposure}}.
\newblock In \emph{Conference on Natural Language Learning}, pages 268--278.

\bibitem[{Dunn and Wong(2022)}]{dunn-wong-2022-stability}
J.~Dunn and S.~Wong. 2022.
\newblock \href {https://aclanthology.org/2022.coling-1.3} {{Stability of
  Syntactic Dialect Classification over Space and Time}}.
\newblock In \emph{Proceedings of the 29th International Conference on
  Computational Linguistics}, pages 26--36.

\bibitem[{Ellis(2007)}]{Ellis2007}
N.~Ellis. 2007.
\newblock \href {https://doi.org/https://doi.org/10.1093/applin/ami038}
  {{Language Acquisition as Rational Contingency Learning}}.
\newblock \emph{Applied Linguistics}, 27(1):1--24.

\bibitem[{Goldberg(2006)}]{g06a}
A.~Goldberg. 2006.
\newblock \emph{{Constructions at Work: The Nature of Generalization in
  Language}}.
\newblock Oxford University Press, Oxford.

\bibitem[{Goldberg(2019)}]{Goldberg2019}
A.~Goldberg. 2019.
\newblock \emph{Explain Me This}.
\newblock Princeton University Press.

\bibitem[{Goldsmith(2001)}]{Goldsmith2001}
J.~Goldsmith. 2001.
\newblock \href {https://doi.org/https://doi.org/10.1162/089120101750300490}
  {{Unsupervised Learning of the Morphology of a Natural Language}}.
\newblock \emph{Computational Linguistics}, 27(2):153--198.

\bibitem[{Goldsmith(2006)}]{Goldsmith2006}
J.~Goldsmith. 2006.
\newblock \href {https://doi.org/https://doi.org/10.1017/S1351324905004055}
  {{An Algorithm for the Unsupervised Learning of Morphology}}.
\newblock \emph{Natural Language Engineering}, 12(4):353--371.

\bibitem[{Grave et~al.(2018)Grave, Bojanowski, Gupta, Joulin, and
  Mikolov}]{Grave2018}
E.~Grave, P.~Bojanowski, P.~Gupta, A.~Joulin, and T.~Mikolov. 2018.
\newblock \href {http://arxiv.org/abs/1802.06893} {{Learning Word Vectors for
  157 Languages}}.
\newblock In \emph{Proceedings of the International Conference on Language
  Resources and Evaluation}, pages 3483--3487.

\bibitem[{Jackendoff(2013)}]{10.1093/oxfordhb/9780195396683.013.0005}
R.~Jackendoff. 2013.
\newblock \href {https://doi.org/10.1093/oxfordhb/9780195396683.013.0005}
  {{Constructions in the Parallel Architecture}}.
\newblock In \emph{{The Oxford Handbook of Construction Grammar}}, pages
  70--92. Oxford University Press.

\bibitem[{Kesarwani(2018)}]{Kesarwani2018}
A.~Kesarwani. 2018.
\newblock \href {https://www.kaggle.com/datasets/aashita/nyt-comments} {{New
  York Times Comments}}.
\newblock Kaggle.

\bibitem[{Langacker(2008)}]{l08}
R.~Langacker. 2008.
\newblock \emph{{Cognitive Grammar: A Basic Introduction}}.
\newblock Oxford University Press, Oxford.

\bibitem[{Nguyen et~al.(2016)Nguyen, Nguyen, Pham, and Pham}]{nn}
Dat Quoca Dai Quocb Dat Quoca Dai~Quocb Nguyen, Dat Quoca Dai Quocb Dat Quoca
  Dai~Quocb Nguyen, Dang~Ducc Pham, and Son~Baod Pham. 2016.
\newblock \href {https://doi.org/10.3233/AIC-150698} {{A Robust
  Transformation-based Learning Approach Using Ripple Down Rules for
  Part-of-Speech Tagging}}.
\newblock \emph{AI Communications}, 29(3):409--422.

\bibitem[{Ortman(2018)}]{Ortman2018}
M.~Ortman. 2018.
\newblock \href
  {https://www.kaggle.com/datasets/mikeortman/wikipedia-sentences} {{Wikipedia
  Sentences}}.
\newblock Kaggle.

\bibitem[{Petrov et~al.(2012)Petrov, Das, and McDonald}]{pdm12}
S.~Petrov, D.~Das, and R.~McDonald. 2012.
\newblock \href
  {http://www.lrec-conf.org/proceedings/lrec2012/pdf/274_Paper.pdf} {{A
  Universal Part-of-Speech Tagset}}.
\newblock In \emph{Proceedings of the Eighth Conference on Language Resources
  and Evaluation}, pages 2089--2096. European Language Resources Association.

\bibitem[{Rae et~al.(2019)Rae, Potapenko, Jayakumar, and Lillicrap}]{Rae2019}
J.~Rae, A.~Potapenko, S.~Jayakumar, and T.~Lillicrap. 2019.
\newblock \href {https://doi.org/10.48550/ARXIV.1911.05507} {{Compressive
  Transformers for Long-Range Sequence Modelling}}.

\bibitem[{Schler et~al.(2006)Schler, Koppel, Argamon, and
  Pennebaker}]{Schler2006}
J.~Schler, M.~Koppel, S.~Argamon, and J.~Pennebaker. 2006.
\newblock \href {http://www.cs.biu.ac.il/~schlerj/schler_springsymp06.pdf}
  {{Effects of Age and Gender on Blogging}}.
\newblock In \emph{Proceedings of 2006 AAAI Spring Symposium on Computational
  Approaches for Analyzing Weblogs}.

\bibitem[{Steels(2017)}]{Steels2017}
L.~Steels. 2017.
\newblock \href {https://doi.org/https://doi.org/10.1075/cf.00002.ste} {{Basics
  of Fluid Construction Grammar}}.
\newblock \emph{Constructions and Frames}, 9(2):178--255.

\bibitem[{Tiedemann(2012)}]{Tiedemann2012a}
J.~Tiedemann. 2012.
\newblock \href {https://aclanthology.org/L12-1246/} {{Parallel Data, Tools and
  Interfaces in {OPUS}}}.
\newblock In \emph{Proceedings of the Eighth International Conference on
  Language Resources and Evaluation}, pages 2214--2218.

\bibitem[{Zeman et~al.(2018)Zeman, Haji{\v{c}}, Popel, Potthast, Straka,
  Ginter, Nivre, and Petrov}]{zeman-EtAl:2018:K18-2}
D.~Zeman, J.~Haji{\v{c}}, M.~Popel, M.~Potthast, M.~Straka, F.~Ginter,
  J.~Nivre, and S.~Petrov. 2018.
\newblock \href {http://www.aclweb.org/anthology/K18-2001} {{{CoNLL} 2018
  Shared Task: Multilingual Parsing from Raw Text to Universal Dependencies}}.
\newblock In \emph{Proceedings of the {CoNLL} 2018 Shared Task: Multilingual
  Parsing from Raw Text to Universal Dependencies}, pages 1--21.

\bibitem[{Zeman et~al.(2017)Zeman, Popel, Straka, Haji{\v{c}}, and
  Others}]{zeman-etal-2017-conll}
D.~Zeman, M.~Popel, M.~Straka, J.~Haji{\v{c}}, and Others. 2017.
\newblock \href {https://doi.org/10.18653/v1/K17-3001} {{C}o{NLL} 2017 {S}hared
  {T}ask: Multilingual parsing from raw text to {U}niversal {D}ependencies}.
\newblock In \emph{Proceedings of the {C}o{NLL} 2017 Shared Task: Multilingual
  Parsing from Raw Text to Universal Dependencies}, pages 1--19.

\bibitem[{Zhang et~al.(2015)Zhang, Zhao, and LeCun}]{Zhang2015}
Xiang Zhang, Junbo Zhao, and Yann LeCun. 2015.
\newblock \href {https://doi.org/10.48550/ARXIV.1509.01626} {Character-level
  convolutional networks for text classification}.

\end{thebibliography}

\end{document}